\documentclass{article}

\usepackage[utf8]{inputenc}
\usepackage{amsmath}
\usepackage{natbib}
\usepackage{amsfonts}
\usepackage{amssymb}
\usepackage{amsthm}
\usepackage{graphicx}
\usepackage{subcaption}
\usepackage{url}
\usepackage{xcolor}
\usepackage{enumitem}

\usepackage[letterpaper,margin=1in]{geometry}

\newcommand{\xhdr}[1]{\vspace{1mm} \noindent\textbf{#1.}}

\newtheorem{remark}{Remark}

\newtheorem{condition}{Condition}

\newtheorem{definition}{Definition}

\newcommand{\reals}{\mathbb{R}}

\newcommand{\vb}{\vec{b}}
\newcommand{\vc}{\vec{c}}

\newcommand{\vs}{\vec{s}}

\newcommand{\vx}{\vec{x}}

\newcommand{\vtheta}{\vec{\theta}}

\newcommand{\cX}{\mathcal{X}}

\newcommand{\cF}{\mathcal{F}}

\usepackage{color-edits}
\addauthor[Hoda]{hh}{red}
\addauthor[Alex]{ajl}{blue}

\author{
   \makebox[.45\textwidth]{Alex John London$^*$}\\
   Carnegie Mellon University\\
   \url{ajlondon@andrew.cmu.edu} \\
   \and
   \makebox[.45\textwidth]{Hoda Heidari\footnote{Authors contributed equally to this work. Alex John London acknowledges support from the U.S. National Science Foundation (IIS2112633). 
Hoda Heidari acknowledges support from NSF (IIS2040929 and IIS2229881)
and PwC (through the Digital Transformation
and Innovation Center at CMU). Any opinions, findings, conclusions, or recommendations expressed in this material are those of the authors and do not reflect the views of NSF or other funding agencies.}}\\
   Carnegie Mellon University\\
   \url{hheidari@andrew.cmu.edu}\\
}

\title{Beneficent Intelligence: 

A Capability Approach to Modeling Benefit, Assistance, and Associated Moral Failures through AI Systems}

\date{}

\begin{document}

\maketitle

\begin{abstract}
The prevailing discourse around AI ethics lacks the language and formalism necessary to capture the diverse ethical concerns that emerge when AI systems interact with individuals. Drawing on Sen and Nussbaum's capability approach, we present a framework formalizing a network of ethical concepts and entitlements necessary for AI systems to confer meaningful \emph{benefit} or \emph{assistance} to stakeholders. Such systems enhance stakeholders' ability to advance their life plans and well-being while upholding their fundamental rights. We characterize two necessary conditions for morally permissible interactions between AI systems and those impacted by their functioning, and two sufficient conditions for realizing the ideal of meaningful benefit.  We then contrast this ideal with several salient failure modes, namely, forms of social interactions that constitute unjustified paternalism, coercion, deception, exploitation and domination. The proliferation of incidents involving AI in high-stakes domains underscores the gravity of these issues and the imperative to take an ethics-led approach to AI systems from their inception.
\end{abstract}

\section{Introduction}

It should be axiomatic that producing some sort of benefit is a necessary condition for the ethical and responsible use of Artificial Intelligence (AI). Without some associated benefit, there would be no way to justify the time, effort, and expense, or harms that cannot be entirely eliminated from developing and implementing an AI system.  Similarly, while some issues of fairness deal with relationships of status or standing, such as equal respect and fair process, a large part of the debate surrounding fairness in AI involves the distribution of the benefits and burdens of a technology.  It is somewhat surprising then that, as frameworks and tools for ethical and responsible AI proliferate \citep{AlgorithmWatch}, roughly half don't mention beneficence or the production or conferral of benefit \citep{jobin2019global}. Those that do, frequently characterize benefits in broad strokes, some of which might accrue to individuals, some to groups, some directly associated with the operation of AI systems and some associated with the economic benefits to developers \citep{jobin2019global, ryan2020artificial}.  At the extreme, some of these forms of benefit---such as the economic reward of selling AI systems---are consistent with the proliferation of AI applications that don't actually perform their stated task and do not provide a direct benefit to anyone impacted by their operation \citep{raji2022fallacy}.  This prospect echoes recent criticisms that frameworks for ethical and responsible AI are too centrally focused on the values and interests of technology creators \citep{birhane2022values} and on the way that relevant moral values are instantiated within technologies rather than on the larger impacts of those technologies \citep{wang2022against, laufer2022four}, including questions of authority and power dynamics, human sovereignty and autonomy, and democratic values like liberty and freedom \citep{washington2020whose,barabas2020studying}.

In contrast, other documents embrace a more demanding ambition of creating AI systems that ``enhance human well-being, empowerment and freedom'' \citep{shahriari2017ieee}. For example, the Asilomar AI Principles hold that ``The goal of AI research should be to create not undirected intelligence, but beneficial intelligence'' and connect benefit with the goal to ``empower as many people as possible'', a phrase also used by the Partnership on AI (\citeyear{PAI}) and the \citet{EC-TAI}.  ``Who is to be empowered and how'' is not always clear, however, and the relationship between empowerment and well-being is largely implicit.  Nevertheless, these views are striving to put, ``human flourishing at the center of IT development efforts'' and to ensure that ``machines [...] serve humans and not the other way around''~\citep{shahriari2017ieee}.

If the ambition of creating AI systems that empower and enhance human well-being and freedom is focused on individuals who interact with that system, then this amounts to the goal of ensuring that in their interactions with individuals, AI systems \emph{meaningfully benefit} people and function as \emph{assistive} technologies. According to the \citeauthor{WHO}, assistive technology is an umbrella term for systems or products that ``maintain or improve an individual’s functioning and independence, thereby promoting their well-being''. It is common to think of assistive technologies as systems designed for people with disabilities or who face other physical, social, or psychological impediments to their ability to function\footnote{Throughout, we follow Sen and interpret the terminology of `functioning' as the beings and doings that the individual may have reason to value.}.  And there is clear interest in developing AI systems specifically designed for the purpose of assisting such individuals~\citep{london2023ethical}. But firms implicitly invoke this same network of ideas when they describe smart devices as ``virtual assistants''~\citep{copilot,chatgpt} or as ``cognitive accessibility assistive technology''~\citep{cognitive_assistance}. In such cases the implicit, if not the explicit, claim is that such AI systems will help users overcome some common human limitation and expand their capabilities so that they can advance their well-being more effectively or efficiently.  

The conception and imagery of assistance through AI is attractive, because assistive interactions are morally valuable and meritorious, they promote perceptions of trustworthiness and reliability toward the technology, and they focus explicitly on the relationship between the functioning of the technology and its impact on the agency and well-being of individuals.  Exploring the requirements for an AI system to be assistive provides an opportunity to distinguish different ways in which AI systems might benefit stakeholders, and provide a clear and precise model of the relationship between individual agency and well-being in a way that highlights neglected issues of power, agency, and autonomy.

\xhdr{Contributions} In what follows, we offer a formal model of assistance that builds on Nussbaum and Sen's capability approach (\citeyear{sen1999commodities,nussbaum2009creating}) (Section~\ref{sec:model}). The primary goal of this model is to articulate and defend precise criteria for assistance with the goal of clarifying the conditions necessary for an AI system to be aligned with the life plans and well-being of its individual users (Section~\ref{sec:constraints}). Our model clarifies the relationship between agency, freedom, and well-being in assistive interactions and allows us to show how common deviations from these values constitute morally problematic social interactions, such as coercion, deception, unjustified paternalism, domination, and exploitation (Section~\ref{sec:failure_modes}). Representing these differences usefully expands the array of ethical concerns that are relevant to assessing decisions across the AI lifecycle of conception, design, development and deployment, and facilitates the ability of stakeholders to evaluate the merits and risks of AI systems and the extent to which they are assistive, beneficial, or empowering.  Because the criteria for success or failure for such systems inherently involves ethical considerations, factors such as the impact of a system on user agency, autonomy, and well-being must be taken into account from the inception of the technology and across the life-cycle of system design, development and use.  As we discuss in Section~\ref{sec:discussion}, we, therefore, hope that a formal model for assistance will facilitate an approach to system development in which ethical values are proactively considered and inform more technical considerations of system functionality and performance in later stages of the AI lifecycle.\footnote{Given the formal nature of the model presented here, it may be tempting to implement it in computer-readable form and automate compliance with the criteria presented here. We state, however, that this is \emph{not} currently the intended use of our proposed framework, and we anticipate significant challenges, both ethical and technical, to such implementation. The goal of our formal model is to bridge the communication gap between experts in AI and ethics, by putting forth a common dictionary that defines and expands salient ethical considerations in developing AI products and systems.}

\subsection{Related Work}\label{sec:related}

\xhdr{Connection to the responsible AI literature} A growing chorus of researchers in academia, the private and public sectors alike have attempted to document, formalize, and remedy the adverse social ramifications of AI systems in high-stakes domains. These efforts have been largely reactive in nature, aiming to restore/promote a handful of values and principles (e.g., safety, privacy, fairness, transparency, and accountability) \emph{after} an AI system has been created with harmful consequences. Increasingly, voices from within the research community have called on AI experts and practitioners to stop treating value alignment as an after-thought and think carefully about how AI products can impact the interests and well-being of marginalized stakeholders~\citep{wang2022against,coston2022validity,passi2019problem}. 
We heed these calls by providing a normative framework that centers the basic rights and well-being of impacted individuals and stakeholders---from the inception of an AI project. In addition to the common principles of Responsible AI, such as privacy, our framework encourages reflection on \emph{individual-level}, \emph{context-specific} values (e.g., accessibility and social affiliation in the context of elderly care), to ensure that AI products and services meaningfully assist stakeholders in advancing their life plans.

\xhdr{Prior work on the capabilities approach to AI} There has been considerable interest in capabilities within ethics \citep{nussbaum2007frontiers} and political philosophy \citep{nussbaum2000women,sen1995inequality,anderson1999point}. Interest in the capabilities approach is grounded in its ability to provide a unified account of the relationship between the \emph{basic rights} of individuals, their \emph{freedom}, and their \emph{well-being}.  This connection is forged from a precise conception of the way that agents convert \emph{resources} into \emph{functionings}, understood as ways of being and doing. (We will formalize these concepts in Section~\ref{sec:model}).  A significant body of work has applied these concepts to issues in bioethics \citep{buchanan2008autonomy, daniels2001justice, ruger2004health, venkatapuram2013health}, research ethics \citep{pratt2015global, london2021common}, disability \citep{brownlee2009disability,kittay2010cognitive} and elsewhere.  Outside of work in economics \citep{sen1997choice,sen1999commodities}, this scholarship tends to focus on philosophical ideas rather than formal models.  The capabilities approach has had limited influence in AI ethics \citep{bakiner2022academics} with some notable exceptions \citep{jacobs2020capability,jacobs2020two,coeckelbergh2011human,coeckelbergh2010health,borenstein2010robot, bondi2021envisioning}. Prior work tends to involve discursive argumentation without a formal model. It also tends to focus on ensuring that basic rights are protected without clarifying the relationship between capabilities, basic rights or entitlements, and individual well-being. 
A central innovation of our work is to provide a formal model that explicitly represents the relationship between basic entitlements, freedom, and individual well-being. On our view, assistance must go beyond rectifying a capability deficit \citep{jacobs2020capability} to ensuring that the deficit that is rectified enables the recipient to pursue projects that contribute to their real freedom or well-being.  This is necessary to avoid interactions undertaken with the intent of providing assistance but that actually constitute a form of  (unjustified) paternalism, domination, or other problematic interactions.

Another important contribution of our work is to expand the portfolio of ethical issues that are relevant to the assessment of AI systems.  A side-effect of the extensive focus on issues of fairness and bias in the literature on ethics in AI is that a wide range of problematic relationships are often grouped under these general terms, limiting the precision of the ethical analysis that is provided.  By articulating a model to explicitly distinguish assistance from justified and unjustified paternalism, coercion, domination and exploitation, we offer a finer partition of ethical dimensions along which AI systems can be evaluated.  

Finally, the model that we articulate can be seen as complimenting approaches, such as capability sensitive design, that provide procedures for engaging with stakeholders in order to elicit information that will facilitate the production of systems that achieve ethically desirable ends \citep{jacobs2020capability}.

\xhdr{Comparison with consequentialist approaches to AI ethics}  It is common in both ethics and AI to represent the well-being of individuals as a utility function.  Although this has numerous operational advantages, it encodes the morally contentious assumptions that every aspect of an individual's life can be mapped onto a single scale of value and that there is nothing morally objectionable, per se, in having to make trade-offs in any aspect of a person's life. This representation is thus incompatible with certain deontological, specifically certain Kantian, views according to which respect for the status of individuals as ends in themselves requires that certain of their interest not be represented as having a dollar value and, thus, as available to trade off in pursuit of material gain \citep{bjorndahl2017kantian}. 

Our approach provides a representation of individual well-being that is rich enough to capture the diversity in goals, projects, and plans that give meaning to the lives of diverse agents while avoiding the implication that all aspects of a person's life can be valued on a single scale.  It builds in the insight foundational to the capabilities approach that individuals have a right to a basic social minimum and that this should be represented as a claim to be able to function above a threshold in distinct spheres or aspects of life.  This basic minimum is necessary for the real freedom of individuals to do and to be.  However, our view also accommodates the insight that, in the course of forming, pursuing, and revising a life plan of one's own, agents often value certain activities or ways of being over others and that acting on such priorities is often essential to the ability to pursue a life plan of one's own. 

\xhdr{Assistance through AI} \citet{wandke2005assistance} explicates assistance in a way that emphasizes the ability of technology to expand the space of goals that agents can achieve.  However, the definition of assistance provided focuses on user access to the functions of a machine and the associated taxonomy of forms of assistance is based on different stages of machine functioning.  \citet{newman2022helping} treat assistance as a perspective from which technologies can be assessed and provide a formal model of assistance as an asymmetric relationship in which a human and a robot seek to bring about a goal set by the human.  However, this model does not explicitly encode information about a user's rights or well-being.  As a result, its definition of assistance is too broad and too narrow at the same time.  It is overly broad in that it can be satisfied in cases where agents pursue goals that are trivial and not meaningfully connected to their well-being, and there are no provisions for ensuring that the means used to advance the desired goal conform to relevant ethical requirements.  It is too narrow in that it applies most directly to robotic systems rather than to AI systems that might work to realize user goals through indirect means, such as by contacting a human service provider and tasking them with an action necessary to assist a human.  
In contrast, our model is more general and explicitly represents information related to the freedom, autonomy, and well-being of the user.  It is thus better able to distinguish cases of trivial benefit from meaningful benefit or true assistance.  Importantly, it is also able to distinguish assistive interactions from cases in which the system and the agent pursue the same goal for reasons that are morally problematic.

\xhdr{Merits of economic modeling} In order to bridge the gap between the qualitative, narrative approach of ethics and the tendency of AI experts and developers for quantitative, precise problem formulations~\citep{selbst2019fairness}, we provide a \emph{mathematical} model of an agent's functionings, capabilities, freedoms, and life-plans with the aim of distilling the key moral dimensions of individuals' interactions with AI systems in a way that is faithful to the underlying normative theory, and simultaneously provides adequate formalization for consumption by an AI audience. 

Mathematical models are powerful tools intended to provide simplified yet precise representations of complex socioeconomic systems~\citep{morgan2012world}. It is inherent to practically useful models to possess a certain degree of simplification~\citep{friedman1953methodology}. This abstraction aims to guide our attention to crucial components and properties of the system. A useful abstraction facilitates structured reasoning, improves understanding of the phenomenon at hand, and allows the researcher to explore, describe, or explain real-world observations~\citep{morgan2008models}.
Compared to narrative explanations of a system, a model ``disciplines [the researcher's] thinking by making [them] specify exactly what [they] mean''~\citep{OpenStax}, and gives structure to the type of inferences they can make~\citep{morgan2012world}. 

According to \citet{morgan2012models}, two of the primary functions of a mathematical model are (1) to ``develop theory, concepts and classificatory systems'' and (2) to ``suggest explanations for certain specific or general phenomena''. In addition to allowing us to precisely define and compare under-studied issues, such as domination, exploitation, and coercion, to which AI systems may contribute, our model offers a new perspective on several recent controversies involving AI products and systems. Finally, we hope the framework proposed here promotes an ethics-led approach to the design and development of future AI systems in high-stakes domains, by priming AI designers and creators to reflect on the ways in which their product can provide meaningful benefit to the intended stakeholders and to anticipate the ``ripple effects''~\citep{selbst2019fairness} of their work on basic rights and well-being of users and impacted individuals.

\begin{table}[t!]
\begin{center}
\begin{tabular}{c|p{0.8\textwidth}}
Notation                 & Description \\ \hline \hline
$\vx_i$        & A vector representing the commodities and resources available to $i$.\\
$\cX_i$        & The set from which $i$ can choose the resource vector ($\vx_i \in \cX_i$). \\
$\vc_i$        & The (constant) vector representing $i$'s characteristics through which they can turn resources into functionings.\\
$\vs_i$        & The (constant) vector of relevant conditions of the society $i$ lives in. \\
$f(\vx_i; \vs_i, \vc_i)$ & A pattern of use $i$ can utilize to turn $\vx_i$ into their choice of functionings given $\vc_i$ and $\vs_i$. \\
$\cF$        & The set of all patterns of use $i$ can choose from (i.e., $i$ can choose $f \in \cF$). \\
$\vb_i$        & A vector of functionings $i$ could obtain by choosing $f \in \cF$: $\vb_i = f (\vx_i; \vs_i, \vc_i)$. \\
$B_i$        & The space representing all imaginable functioning vectors for $i$ that are relevant to some plausible life plan $i$ might adopt ($\vb_i \in B_i$).\\
$Q_i$          & Agent $i$'s freedom consisting of all functioning vectors available to them through the choice of $\vx_i \in \cX_i$ and $f \in \cF$. 
\end{tabular}
\caption{A summary of key concepts and the corresponding notations introduced in Section~\ref{sec:model}.} 
\label{tab:summary_2}
\end{center}
\end{table}

\section{A Model of Agents' Functionings and Freedom}\label{sec:model}
In this section, we offer a formal model of agent's relevant functionings and freedoms, so that we can articulate and defend precise criteria for assistance through AI.  For an AI system to be assistive, it must function in a way that enables the recipient of assistance to better advance the plans and projects that matter to them.  Here, we build on Sen's notation~\citep{sen1999commodities} for representing the factors that influence an individual's ability to function, how that ability relates to freedom, and how the choice to pursue certain activities, from a larger field of possibilities, relates to that individual's conception of a good life. 
With these building blocks n place, in the following sections, we define the criteria for AI to constitute a beneficent and assistive system and we show how deviations from these criteria constitute one or more types of ethically problematic interactions. 

We introduce the notation necessary to represent the freedoms of an individual agent $i$\footnote{Note that throughout this work, an agent refers to a humanbeing, \emph{not} an AI agent.} who is potentially impacted by an AI technology\footnote{The formalism and arguments developed in this work are applicable to individuals meaningfully impacted by AI systems even if they are not the direct \emph{users} of the technology. Examples include an older adult whose children purchase and install a surveillance device in her home. Another example is a job-seeker whose employment prospects are impacted by a resume-screening software that automatically filters their application in/out of further consideration. In these examples, the impacted individual is not a user of the system and they may even be unaware of the system's existence, nonetheless, they are impacted by it in significant ways.}. For example, $i$ may be an older adult who interacts with a smart virtual assistance device to facilitate their day-to-day activities, such as reading or grocery shopping. As another example, $i$ can be a job seeker who interacts with an AI-powered video interview software.

\xhdr{Commodities and resources} Let $\vx_i$  denote the vector of commodities and resources $i$ can access. As concrete examples in the context of caring for older adults, different components of $\vx_i$ might represent $i$'s housing circumstances, monetary wealth, the reading material they own or that is available in local libraries. We treat an AI system as a commodity $i$ can access, if, for example, it is available for purchase. Generally, we also treat information, such as how to use an AI system or the availability of job opportunities, as a resource. 
    
Let $\cX_i$ be the set of all commodity vectors that are practically accessible to $i$.  A vector $\vx'_i$ is accessible to agent $i$ possessing initial commodity vector $\vx_i$ if $i$ can convert $\vx_i$ into $\vx'_i$---e.g., the agent could convert some of their money into a new commodity by purchasing it, or they could use their resources to access an institution through which new resources can be attained, such as accessing books through a library. 
    
\xhdr{Agent's characteristics} Let the vector $\vc_i$ refer to characteristics of $i$ in virtue of which they are able to convert a vector of available resources, $\vx_i$, into a functioning.\footnote{In Nussbaum's terminology~\citep{nussbaum1999women}, $\vc_i$ captures $i$'s \emph{basic} and \emph{internal} capabilities at the relevant snapshot of time at which we are modeling agents' functionings and capabilities.} One such characteristic includes current physical abilities.  For example, strong legs is one feature of $i$ in virtue of which they can convert a two story house into functionings that constitute safe shelter. As this feature of $i$ changes over time (e.g., an older adult physically weakens), their ability to function safely within a two story house may similarly become impaired.  Likewise, the physical ability to see is a characteristic that enables reading. But the ability to read draws heavily on a second attribute of $\vc_i$, namely, the acquired skill that includes knowledge of a language (e.g., English), the written alphabet, and how to use that knowledge to extract information encoded in books printed in that language. 

\xhdr{Social conditions} We depart from \cite{sen1999commodities} by introducing $\vs_i$ to refer to a vector of relevant properties of the society in which $i$ lives. One set of features represented in $\vs_i$ may have to do with infrastructure and the kinds of functionings that it can support.  In a society where libraries are only located in major population centers and in which roads are poor and mass transit does not extend to rural areas, people who live in rural areas may have to spend too much time traveling to feasibly access books. Other components of $\vs_i$ can represent the social norms in force in the society in which $i$ lives.  For example, different societies might regard specific types of attire as acceptable for various careers.  Whether $i$ can find and retain gainful employment can depend on whether they have sufficient resources to show up to work in ways that conform to these social expectations.

\xhdr{Functionings}  
Let $f$ be a vector-valued utilization function characterizing one \emph{pattern of use} that $i$ can enact. In particular, $f$ maps the society $i$ lives in ($\vs_i$), $i$'s characteristics ($\vc_i$), and their commodity vector ($\vx_i$), to a vector of beings and doings (or functionings) that $i$ can achieve, denoted by $\vb_i = f(\vx_i; \vs_i, \vc_i)$ . Within this framework, reading her choice of books is a functioning that may be \emph{available} to an older adult $i$, and captured as a part of her functioning vector, $\vb_i$, if she chooses the appropriate pattern of use (i.e., the appropriate $f$), has the developed skill to read the English language, has the physical ability and the resources necessary to travel to the local library to check out books, and lives in a society where there are libraries and in which $i$ is not excluded by a social norm from traveling to the library or checking out books. We will denote by $F$ the set of all patterns of use that are available to the agent.

It is important to note that, at the time of choice, $i$ can choose $f \in F$ and $\vx_i \in \cX_i$, but they cannot choose their $\vc_i$ and $\vs_i$. For our purposes, we thus treat $\vc_i$ and $\vs_i$ as fixed and given.  However, over longer time horizons, both $\vc_i$ and $\vs_i$ can be influenced by the actions of the agent.  With regard to $\vc_i$, Aristotle noted that agents develop certain traits or characteristics by first acting in the relevant way \citep{burnyeat1980aristotle,london2001moral}.  For example, adult learners have to be taught to speak and read a second language. This requires certain resources (someone to act as a teacher, educational materials suited to the capacity they will use to extract information from symbolic encoding, such as written texts or Braille texts), and then practice at putting letters together to form words.  In this case, we say the student chooses to enact a pattern of use $f$ which, when repeated over time, develops a feature of the student, namely, the skill of reading, captured as a component of $\vc_i$. In contrast, an older adult's physical and mental decline can alter certain elements of $\vc_i$, hindering their ability to do things they were normally able to do.  Likewise, an agent $i$ may be able to influence $\vs_i$ by engaging in protest or social reform or by moving to a different community.

\xhdr{Freedoms} We assume functioning vectors (i.e., $\vb_i$'s) belong to a space $B_i$, which represents all of the 
ways of being that are available to $i$ and are relevant to some plausible life plan $i$ might adopt.  Sen represents $i$'s freedom to choose functionings with $Q_i$, defined as the set of functionings $i$ can achieve given their command of resources (captured by $\cX_i$), and their ability to convert resources into functionings in light of their characteristics, $\vc_i$.  We add to this the social conditions $i$ lives in, $\vs_i$.  More precisely, this yields a definition of the agent's freedom as: $$Q_i = \Bigl\{ \vb_i \in B_i \vert \exists f \in F \textbf{ }\exists \vx_i \in \cX_i \left(\vb_i = f (\vx_i; \vs_i, \vc_i) \right) \Bigr\}.$$ 
However, for reasons articulated in \citep{sugden2003opportunity}, we resist defining freedom directly on this space.  Instead, in the next section, we define two perspectives from which the \emph{real} freedom of an agent can be assessed. 

It is important to note that the notions of freedom we define in this section and the next all refer to the set of options the agent has \emph{ex-ante}---i.e., prior to choosing to realize a specific functioning vector. The framework offered here should, therefore, be understood as evaluating interactions according to their impact on the agent's freedoms to function. The agent ultimately has to choose a single functioning vector, but our normative criteria are not applicable \emph{ex-post}---after the agent realizes a specific functioning vector.

Table~\ref{tab:summary_2} summarizes the notations and constructs introduced in Section \ref{sec:model}. 

Before turning to the next section, two remarks are in order. 
First, throughout and for simplicity, we focus on adult agents who are competent to make their own decisions. The question of ``who are the agents whose interactions must be accounted for'' is left largely open here, due to the heavily context-depended nature of this choice. In broad strokes, we believe priority must be given to impacted individuals and communities who have been historically under-served and underprivileged. 

Second, our model only attempts to capture the relevant functioning, capabilities, and freedoms of the agent at the time of their interaction with the AI system---not the dynamic interaction of actions and capabilities across an entire life. The choice of the time span relevant to the interaction is itself ethically significant, but we leave the discussion of those considerations out of this contribution. In sections \ref{sec:constraints}--\ref{sec:failure_modes}, we consider the implications of this model for a wider range of stakeholders whose conduct influences one or more parameters necessary for ensuring successful assistance to $i$ and avoiding common modes of failure.

\section{Two Ethical Entitlements}\label{sec:constraints}
Next, we use the model introduced in the previous section to represent two fundamental entitlements of individuals.  First is the moral entitlement to a set of basic capabilities necessary to have the real freedom to formulate, pursue, and revise a life plan. Second is the moral claim to be free to pursue any reasonable life plan of one's choice as long as it does not restrict or impinge on the basic entitlements of others.  The latter moral claim requires a representation of individual well-being conceived of as a life plan whose enactment constitutes the good life for the individual.

 \subsection{Guaranteeing Basic Rights, Liberties and Social Conditions} 
In a pluralistic society, individuals are likely to disagree about what constitutes a good life.  In the face of such pluralism and disagreement, several scholars have argued that there is a higher-order perspective from which such agents can see themselves as sharing the same basic interest in having the real freedom to be able to formulate, pursue and revise a life plan of their own \citep{rawls1971atheory,london2021common}.  To ensure that all agents have the real freedom to pursue a distinctive life plan of their own, every individual is entitled to a basic set of rights, liberties and social conditions that constitute the all-purpose means necessary to be able to form, pursue, and revise a life plan. For example, \citet{rawls1971atheory} argues that all persons are entitled to a basic set of primary goods consisting of ``rights and liberties, opportunities and powers, income and wealth'' (p. 92).\footnote{For an overview of how philosophers from differing traditions express a similar set of views see \citep{london2021common} p. 140-146.} This emphasis on rights and liberties is echoed in policy documents such as the Universal Declaration of Human Rights \citep{assembly1948universal}.  We follow Sen (\citeyear{sen1997choice,sen1995inequality}) and Nussbaum (\citeyear{nussbaum2000women,nussbaum2007frontiers}) in holding that this basic set of entitlements is best represented as a set of fundamental capabilities. These capabilities are $fundamental$ in the sense that they constitute building blocks---rudimentary abilities---on which agents rely to formulate, pursue and revise some conception of the good life. \citet{nussbaum2000women} identifies 10 ``central human functional capabilities'' or 10 ways of being and doing to which all persons have a right.  This list includes ``life'', ``bodily health'', ``bodily integrity'', ``senses, imagination and thought'', ``emotions'', ``practical reason'', ``affiliation'', ``other species'', ``play'', and ``control over one's environment''. We refer the reader to \citep{nussbaum2000women} for a more detailed description of these ten capabilities.  

We use $E$ to denote the space capturing the agent's level of access to the set of \emph{fundamental} capabilities (for instance, if we use real numbers to rate the agent's access to Nussbaum's ten central capabilities, $E$ corresponds to $\reals^{10}$). We represent the basic minimums to which individuals are entitled with a threshold vector, $\vtheta_i \in E$, such that $ \vtheta_i = \theta(\vc_i)$, where $\vtheta(\vc_i)$ indicates the minimum acceptable level of each of these fundamental capabilities for an agent with characteristics $\vc_i$.\footnote{The dependency of $\vtheta_i$ on the agent $i$ is only due to variations in $\vc_i$. Beyond that, the mapping $\theta$ represents a \emph{universal} minimum set of entitlements, independent of the society $i$ lives in ($\vs_i$) and their resources ($\vx_i$). The dependence on $\vc_i$ is necessary to capture scenarios in which an agent might experience such profound disability that it would be impossible for them to attain minimum levels of certain capabilities, such as control over their environment, that are achievable by agents with less severe disabilities.} We use $Q^*_i$ to denote those ways of doing and being that $i$ has the freedom to choose which satisfy these basic minimums: 
$$Q^*_i = \Bigl\{ \vb_i \in B_i \vert \exists f \in F \text{ } \exists \vx_i \in \cX_i \left( \vb_i = f(\vx_i; \vs_i, \vc_i) \land r(\vb_i) \succeq \vtheta_i \right) \Bigr\},$$
where $r: B_i \rightarrow E$ is a function that maps the agent's choice of functioning vector to their degree of access to basic entitlements. We will use the terminology of \emph{real} freedoms to refer to $Q^*_i$ and distinguish it from $Q_i$---the agent's freedoms, unqualified.

The central moral idea here is that every agent $i$ is entitled to a $Q^*_i$ that is non-empty. When $Q^*_i$ is non-empty, $i$ has the real freedom to function in ways that bring them above the threshold of the basic minimum.  This entitlement is not grounded in any agent's conception of the good life.  Rather, it is grounded in the higher-order interest that all individuals share in being able to formulate, pursue, and revise a life plan of their own and in the idea that being able to function above the minimum threshold on basic capabilities is a social prerequisite for advancing this higher-order interest. 

The claim that every agent $i$ is entitled to a $Q^*_i$ that is nonempty grounds our first necessary condition for an interaction with agent $i$ to be morally permissible, that is, the interaction must not empty $Q^*_i$. By an ``interaction'' with agent $i$, we refer to any way that the actions of another agent or an AI system modifies $\cX_i$, $\vc_i$, or $\vs_i$.  

\begin{condition}[First Necessary Condition for Morally Permissible Interactions]\label{def:condition1}
    Consider an agent $i$ with initial real freedom, $Q^*_i \neq \emptyset$.\footnote{Note that this definition does not address cases in which $Q^*_i$ is empty to begin with--before the interaction. In such cases, a minimal moral obligation is to ensure that $i$'s access to basic capabilities is not further impeded as a result of the interaction.}
    Let $Q^{*'}_i$ denote $i$'s real freedom \emph{following} an interaction.
    A necessary condition for the moral permissibility of the interaction is  
$$Q^{*'}_i \neq \emptyset.$$
\end{condition}

An interaction that violates  Condition~\ref{def:condition1} is a serious moral wrong, setting back an agent's basic interests---something \citet{nussbaum1999women} classifies as a violation of a human right.  Interactions that result in direct physical, social and psychological harms are likely to violate the above condition because they will drop $i$ below at least one relevant component of $\vtheta_i$. That said, Condition~\ref{def:condition1} is relatively weak. In particular, it requires keeping $Q^*_i$ non-empty rather than preserving its size or characterizing how far above the relevant thresholds $i$ must be able to function. This is for two reasons.  First, note that the entitlement to real freedom represents a social minimum---a baseline entitlement---and not a measure of the agent's well-being according to their subjective value structure. Therefore, interactions that reduce the size of $Q^*_i$ may not be morally wrong as long as $i$ can still function above the relevant thresholds and they receive some offsetting benefit in return for ceding some elements of $Q^*_i$.  Second, condition~\ref{def:condition1} is \emph{necessary but not sufficient} for moral permissibility, so interactions that reduce $Q^*_i$ without dropping agent $i$ below any relevant threshold in $\vtheta_i$ can still be morally impermissible if they result in a restriction on the agent's considered life plans---a condition explicated next in Section~\ref{sec:entitlement2}.

\subsection{Respect for Individual Life Plans}\label{sec:entitlement2}
Because $Q^*_i$ represents the freedom of individuals to function in ways that are critical to their ability to formulate, pursue and revise a life plan, it doesn't measure how well a person's life is going from their point of view.  Although every agent shares an interest in having the freedom that $Q^*_i$ represents, different individuals acting on that freedom will develop different goals, values, and ideals that define what for them constitutes a good life.  For example, a writer might value quiet time alone in a climate controlled environment in which they can spend long hours working at a computer.  In contrast, another person may view their job as simply a way to make money while they find their greatest enjoyment spending time in the wilderness, camping and fishing.  The same set of choices might be available to both agents, but their values and goals lead them to evaluate those alternatives in very different ways and to make very different choices, accordingly.

\begin{figure}[t!]
    \centering
    \includegraphics[width=0.7\textwidth]{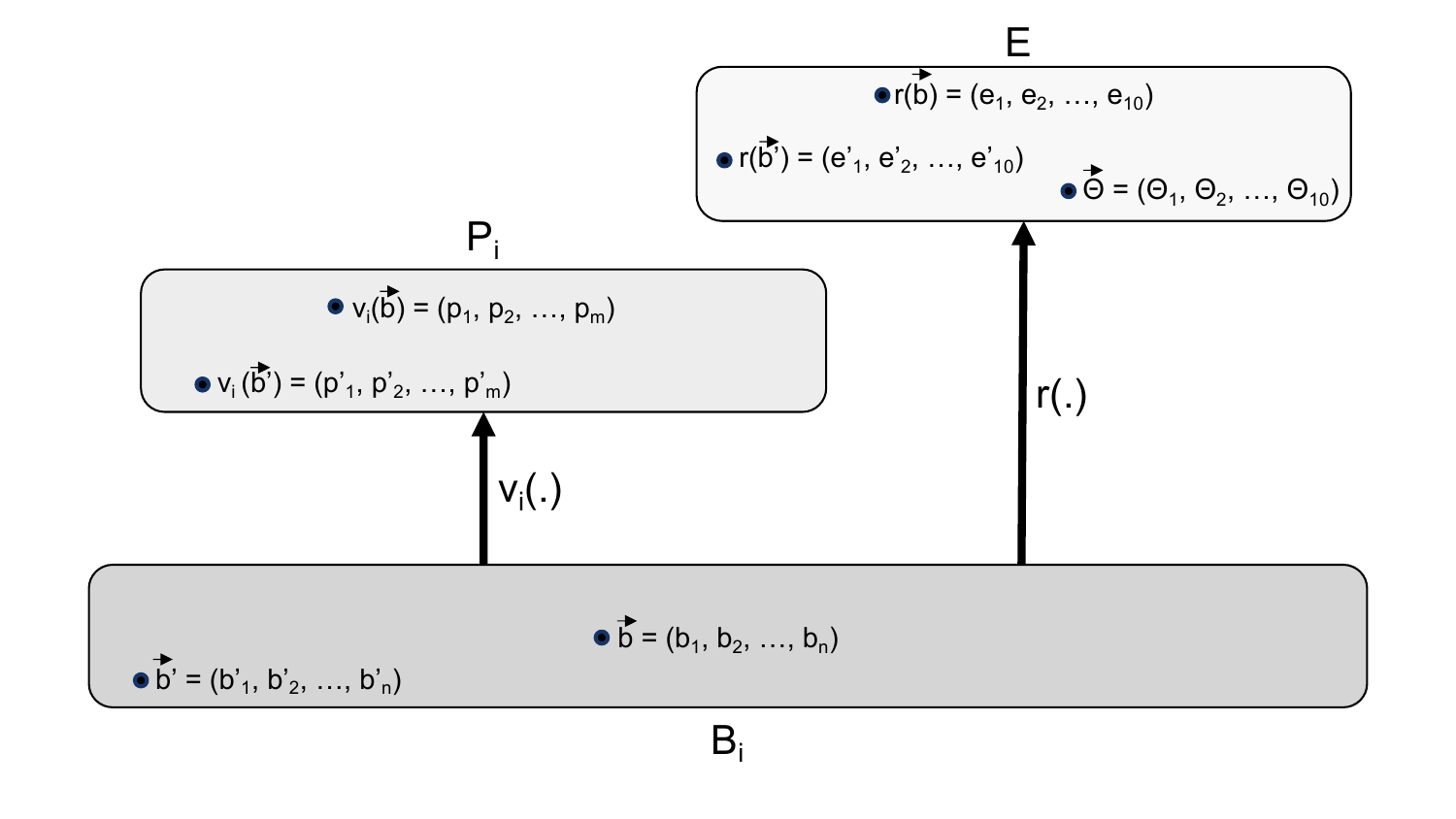}
    \caption{An illustration of the relationship between spaces $B_i, P_i, E$ along with instances belonging to each and how functions $v_i$ and $r$ map these spaces to one another.}
    \label{fig:spaces}
\end{figure}

The agent's values, ideals, and goals together define a life plan for them \citep{london2021common}.  To capture the relationship between these concepts precisely, we define $P_i$ to be the space representing
how well agent $i$ is doing/being with respect to their subjective/individual life goals. (As a concrete example, suppose how well agent $i$'s is doing can be summarized\footnote{This example is provided only for sake of concreteness. We are not advocating for crude approximations of life plans in terms of scalar ratings across only two dimensions.} into a numerical rating of their career achievements, and a numerical rating of their family life. Then $P_i$ corresponds to $\reals^2$). Let $v_i: B_i \rightarrow P_i$ denotes a vector-valued valuation function\footnote{We note that there is precedent for breaking down well-being into multiple dimensions. For example, \cite{alkire2005valuing,narayan2000voices} propose six key dimensions of value: material, bodily, social, and psychological well-being, as well as security and freedom of choice.} mapping every functioning vector, $\vb_i \in B_i$, to a vector of valuations, $v_i(\vb_i)$, where each component of $v_i(\vb_i)$, captures how well $\vb_i$ contributes to each one of the agent's life goals. Note that $v_i(\vb_i)$ represent the value of $\vb_i$ either as a way of realizing some goal/end that $i$ values as constitutive of a good life (e.g., the value of fishing or writing as an activity the agent pursues as an intrinsic goal) or as a means to such a goal or end (e.g., the value of having time off of work as an opportunity to connect with nature). 
For example, for one agent camping and fishing may be a core constituent of a good life as an end in itself, but for another, it may be one, among many ways, of satisfying their goal of connecting with nature. For the former agent, their access to fishing is a component of $P_i$, while for the latter, it is only a component of $B_i$ but not $P_i$. 
Figure~\ref{fig:spaces} illustrates the relationships between $E$, $P_i$, and $B_i$.

\begin{remark}[Distinguishing agent's values, $v_i$, from their mundane preferences, $u_i$]
    Note that $v_i$ reflects agent $i$'s well-informed, considered life plans---that is, it captures how the agent compares different functioning vectors according to how well they fulfill their conception of the good life. When facing particularly difficult choices, agents may have to reflect explicitly about what the values in this structure are and how they relate to choice. More frequently, agents will simply evaluate choices relative to a set of preferences developed over time, which are largely consistent with $v_i$.  In contrast, agents may sometimes make choices that do not reflect $v_i$. For instance, the agent might have a strong desire and so prefer one option over another relative to that desire without considering its implications for their considered values.  In such cases, we say that $i$ chooses relative to another set of preferences.  For our purposes, we use $u_i$ to represent any preference other than $v_i$ that the agent may use to compare different functioning vectors.  When choices are trivial, such as whether to use a mechanical or a wooden pencil to sign a form, choosing relative to $u_i$ might be unproblematic. But, when choices are more momentous, choosing relative to a $u_i$ that diverges from $v_i$ represents a breakdown in agency for $i$.  
\end{remark}

Note that our formulation of the agent's value structure deviates from Sen's formalization, which defined $v_i$ as a \emph{scalar} function mapping functionings to a single value. This deviation is to ensure that (a) $v_i$ can capture multiple dimensions of valuation (e.g., those corresponding to different aspects of flourishing across multiple dimensions of life); and (b) we do not assume a total ordering among functioning vectors. Our representation is more general than Sen's in that it can accommodate any agent whose values can be represented as a single number while also accommodating value pluralists who hold that their values cannot be represented by a single number (see, for example \citep{levi1990hard}) and Kantians who regard representing the worth of agents in terms of the worth of things as a moral transgression \citep{bjorndahl2017kantian}.

Agent $i$ chooses which functionings to realize by evaluating alternative courses of action (i.e., alternative choices of resource vectors, $\vx$'s in $\cX_i$, and alternative use patterns, $f$'s in $F$) relative to this value structure.  Since $v_i$ is vector-valued, it does not necessarily introduce a total ordering over elements of $Q_i$. As such, there is no well-defined notion of a choice function \emph{optimizing} utility. The value structure, $v_i$, however, imposes a partial ordering among functioning vectors, in that $\vb_i$ is considered better than $\vb'_i$ if $v_i(\vb_i) \succ v_i(\vb'_i)$---that is, $\vb_i$ Pareto-dominates $\vb'_i$.\footnote{Given two vectors $\vx = (x_1, \cdots, x_m)$ and $\vx'= (x'_1, \cdots, x'_m)$ both belonging to the same finite-dimensional vector space, say $\reals^m$, $\vx \succeq \vx'$ is equivalent to $\forall i \in \{1,2,\cdots, m\} \left(x_i \geq x'_i\right)$.  In contrast, $\vx \succ \vx'$ is equivalent to $\forall i \in \{1,2,\cdots, m\} \left(x_i \geq x'_i\right) \land \exists j \in \{1,2,\cdots, m\} \left(x_j > x'_j\right).$}
\begin{definition}[Maximal Life Plans]\label{def:m-set}
    Consider an agent $i$ with the (incomplete) preference relation $v_i: B_i \rightarrow P_i$. The \emph{maximal achievable} lives for $i$ is the set of functioning vectors that no alternative in $Q_i$ strictly dominates, that is,
$$M(Q_i, v_i) = \Bigl\{ \vb \in Q_i \vert \nexists \vb' \in Q_i \left(v_i(\vb') \succ v_i(\vb) \right) \Bigr\}.$$
\end{definition}
This set constitutes the life plans that the agent may have reason to pursue in light of their life conditions and values. 

In most cases, the vector of functionings that constitute the maximal achievable lives for adult agents, fortunate enough not to live under conditions of poverty or severe oppression, will also constitute specific ways in which that agent is able to function above the threshold of their basic entitlements.  More precisely the following conditions holds in such cases: 
$$M(Q_i, v_i) \subseteq Q^*_i.$$ 
The reason for this overlap is the fundamental nature of the capabilities represented in $E$.  The life plans of agents are often constructed in a way that both involves the exercise of their basic capabilities and maintains their access to those abilities.  Losing the ability to function above one or more of the thresholds on fundamental capabilities would often hinder or restrict not only a person's real freedoms, $Q^*_i$, but would likely render inaccessible the set of life plans, $M(Q_i, v_i)$, they regard as best relative to their values.
It is important to note, however, that this overlap is not always the case, that is, it is possible for the agent to value a life plan that brings them below the minimum level of some basic capabilities. In such case, the following may hold:
$$M(Q_i, v_i) \setminus Q^*_i \neq \emptyset.$$  In particular, there will almost always be agents whose $Q^*_i$ is non-empty but who nevertheless choose to function in ways that fall below a relevant threshold.  It is a strength of our model that it can formulate conflicts between the social interest in ensuring that community members have real freedom to function in ways that do not bring them below a basic minimum, and instances in which individuals regard such functioning as central to their conception of the good life.  

Whereas respect for the autonomy and agency of individuals entails promoting their ability to function in certain basic respects, it also entails limits on the extent to which society can force agents to actually choose to function in all of the ways necessary to rise above each of these thresholds.  For example, $i$ can have the real freedom to take time off from work and enjoy their hobbies but choose to work longer hours than is socially required because they find personal fulfillment in their job. In this case, respect for $i$'s life plan militates against forcing $i$ to take time off of work.  In contrast, some agents might prefer social activities that are so dangerous (e.g, climbing steep mountains without safety gear) that society can reasonably take steps to prohibit them. When such a prohibition is grounded in concern for the well-being of the individuals whose activities are prohibited, the justification for this policy will hinge on whether the criteria for justified paternalism are met.  We discuss this in Section~\ref{sec:justified_paternalism}.

For present purposes, we want to distinguish cases in which agents might choose to function in ways that fall below a relevant threshold against a background of freedom to do otherwise, from cases in which agents are forced to function below such thresholds. This is the case, for instance, if agent $i$ faces long work hours or a dangerous environment without having the real freedom to relax or to live in safety.  In the latter cases, $i$'s living conditions fall below an acceptable social minimum and so constitute a moral shortcoming to be rectified.  In the former case, they reflect $i$'s distinctive conception of the good life.

Generally speaking, respect for individuals extends to their freedom to decide which functioning vector to realize as long as this does not infringe on the ability of others to develop and exercise their fundamental capabilities (which would violate Condition 1).  As a corollary of this, one important way that an interaction may fail to respect $i$ is by inhibiting $i$'s ability to choose a $b_i$ from their set of maximal lives, $M(Q_i, v_i)$. This is a form of disrespect because such interactions fail to treat $i$ is an agent---the subject of a life that they shape through their choices and that can go better or worse for them. As a weak condition of respect, interactions with agents should not infringe on or restrict their set of maximal life plans.  More precisely:

\begin{condition}[Second Necessary Condition for Morally Permissible Interactions]\label{def:condition2}
    Consider an agent $i$ with initial freedom, $Q_i$.
    Let $Q'_i$ denote $i$'s freedom \emph{following} an interaction.
    The interaction is morally permissible only if
    $$\forall \vb \in M(Q_i, v_i) \textbf{ } \exists \vb' \in Q'_i \left(v_i(\vb')  \succeq v_i(\vb) \right).$$
\end{condition}

This condition reflects the fact that it is $i$'s prerogative to decide among possible life plans in their $M$-set and to choose the functionings that best realize or express their conception of the good life. A morally permissible interaction may shrink $Q_i$, but it should not reduce $i$'s ability to function in ways that could be constitutive of their conception of the good life, or necessary for their ability to function in these ways.  
As we noted in the previous section, interactions that restrict an agent's ability to function relative to basic capabilities in $Q^*_i$ might not violate Condition~\ref{def:condition1} as long as $i$ retains the ability to function above the relevant $\vtheta_i$ threshold. But such interactions are likely to violate Condition~\ref{def:condition2} if the agent relies on those basic capabilities to function in ways that are constitutive of or instrumental for some conception of the good life for them.

Table~\ref{tab:summary_3} summarizes the notations and constructs introduced in Section  \ref{sec:constraints}. 

\begin{table}[t!]
\begin{center}
\begin{tabular}{c|p{0.8\textwidth}}
Notation                 & Description \\ \hline \hline
$\vtheta_i$        & The vector representing the minimum acceptable level of $i$'s access to basic capabilities.\\
$E$        & The space representing level of access to basic capabilities ($\vtheta_i \in E$).\\
$Q^*_i$          & Agent $i$'s real freedom consisting of all available functioning vectors that guarantees $i$ at least a $\vtheta_i$ level of access to all basic capabilities. \\
$P_i$        & The space representing how well $i$ is doing/being according to their conception of the good life.\\
$v_i(\vb_i)$        & Agent $i$'s subjective valuation function mapping a functioning vector, $\vb_i$, to a vector-valued utility function.\\
$r(\vb_i)$        & An objective valuation function mapping a functioning vector, $\vb_i$, to $i$'s level of access to basic capabilities.\\
$M(Q_i, v_i)$        & Maximal life plans for agent $i$.\\
\end{tabular}
\caption{A summary of our key concepts and the corresponding notation introduced in Section~\ref{sec:constraints}.} 
\label{tab:summary_3}
\end{center}
\end{table}

\section{Senses of Beneficence: From Trivial to Meaningful Benefit}

Within the model outlined above, there are multiple senses in which an AI system might confer ``benefit'' to an individual or be \emph{beneficent}. Any interaction with an AI system that \emph{improves}  $i$'s freedoms ($Q_i$), real freedoms ($Q^*_i$), or maximal life plans ($M(Q_i, v_i)$) can be seen as a benefit. More precisely, we say an interaction improves a set $S \subset B$ to $S' \subset B$ with respect to a valuation function $w: B \rightarrow \reals^m$ if and only if
\begin{eqnarray*}
&& \forall \vb \in S \textbf{ } \exists \vb' \in S' \left(w(\vb') \succeq w(\vb) \right) \\
&\land & \exists \vb \in S \textbf{ } \exists \vb' \in S' \left(w(\vb') \succ w(\vb) \right).
\end{eqnarray*}
In other words, the interaction either expands the set $S$, or if it removes an element from it, it replaces it with another one strictly preferred through $w$. 
With this definition in place, we define beneficence as follows: 
\begin{definition}[Beneficence]
    Consider an agent $i$ with initial $Q_i$, $Q^*_i$, and $M(Q_i, v_i)$.
    Let $Q'_i$, $Q'^*_i$, and $M(Q'_i, v_i)$ denote these sets \emph{following} an interaction. The interaction is \emph{beneficial} if at least one of the following conditions holds:
    \begin{enumerate}
        \item[(1)] It improves $Q_i$ to $Q'_i$ with respect to $u_i(.)$. 
        \item[(2)] It improves $Q^*_i$ to $Q'^*_i$ with respect to $r(.)$.
        \item[(3)] It improves $M(Q_i,v_i)$ to $M(Q'_i, v_i)$ with respect to $v_i(.)$.
    \end{enumerate}
\end{definition}
If an interaction only satisfies (1) above, we say that it is only \emph{weakly} beneficial because benefits in this space can include the ability to do things that are trivial or of little importance for the agent. To confer meaningful benefit to $i$, an interaction must satisfy (2) or (3). The reason we refer to these notions of beneficence as meaningful is that (2) concerns the basic capabilities to which all persons are entitled insofar as they share the higher-order interest in being able to formulate, pursue and revise a life plan and (3) concerns the ability of the agent to take advantage of those opportunities in pursuit of the goals and ends that are constitutive of their conception of the good life. In what follows, we further explicate these notions of beneficence.

\subsection{The Weak Sense of Benefit}
The first sense of benefit, on its own, is incredibly weak.  Anything that produces a positive change for $i$'s  $\vx_i$, $\vc_i$, or even $\vs_i$ can be seen as a benefit since it is likely to add a new $\vb_i$ to $Q_i$ expanding the agent's freedom and allowing them to function in a new way. Interactions with an AI provide a benefit in this weak sense if there is some respect in which they improve the agent's ability to do some activity. 
More precisely, 
\begin{definition}[Weak Benefit]
    Consider an agent $i$ with initial freedom, $Q_i$.
    Let $Q'_i$ denote $i$'s freedom \emph{following} an interaction. The interaction is \emph{weakly beneficial} if 
\begin{eqnarray*}
&& \forall \vb \in Q_i \textbf{ } \exists \vb' \in Q'_i \left(u_i(\vb') \succeq u_i(\vb) \right) \\
&\land & \exists \vb \in Q_i \textbf{ } \exists \vb' \in Q'_i \left(u_i(\vb') \succ u_i(\vb) \right).
\end{eqnarray*}
\end{definition}

\citet{danaher2023mechanisms} argue that we increase the agent's moral freedom as soon as we expand the set of options from which they choose their functionings.  However, adding a new $\vb_i$ to $Q_i$ can result in benefits that are utterly trivial and unimportant in the sense that they do not make a person’s life go better in any \emph{meaningful} way---they don't change their access to any basic capabilities or contribute in a strictly positive manner to the projects, plans, or ways of functioning that make up their conception of a good life. More precisely, an interaction can expand $i$'s freedoms (i.e., $Q_i \subset Q_i'$) without improving $Q_i^*$ or any aspect of the agent's maximal life plans (i.e., $M(Q_i', v_i) = M(Q_i, v_i)$)\footnote{For present purposes we treat $v_i$ as fixed and leave relaxation of this assumption to subsequent work.  However, see \citep{pettigrew2015transformative, pettigrew2019choosing} for a discussion of changes to the agent's value structure.}.  We, therefore, regard this notion of benefit as too weak to count as expanding the agent's \emph{real} freedom---their freedom to be and to do in ways that are morally important.

Additionally, this sense of benefit is so weak that there will almost always be some respect in which an AI system can satisfy this condition for some stakeholder agent (e.g., a user willing to purchase the AI product knowing its capabilities).  This, in turn, creates room for unintentional or intentional confusion to the extent that those who would use or otherwise be impacted by an AI system believe that expected or promised benefits will be more meaningful than they are. For example, a voice-controlled smart speaker might allow the user to listen to music without having to first access their phone or their computer. This constitutes a weak benefit in that it enables the user to function in a way they could not before.  Such a benefit might be sufficient for users who enjoy the novelty of new technology or who want to listen to music while they engage in tasks that occupy their hands.  But for many users, this relatively minor convenience will not expand or improve their ability to function in ways that advance their life plans.

\subsection{Meaningful Benefit and Assistance }

 As we use the term, assistance picks out a subset of benefits, namely, those that are meaningful in that they better enable an agent to advance the projects and plans that contribute to or that are constitutive of their conception of the good life. Meaningful benefit can be defined in two ways.

An interaction can confer meaningful benefit to $i$ if it improves the degree of real freedom that $i$ enjoys by expanding or improving $i$'s ability to function relative to one or more of the central capabilities that define $Q_i^*$.    

\begin{definition}[Assistance through Improving Real Freedom]\label{def:assistance_objective}
    Consider an agent $i$ with initial real freedom, $Q^*_i$.
    Let $Q'^*_i$ denote $i$'s real freedom \emph{following} an interaction. The interaction is \emph{beneficial/assistive} through improving $i$'s real freedoms if
\begin{eqnarray*}
    && \forall \vb \in Q^*_i \textbf{ } \exists \vb' \in Q'^*_i \left(r(\vb') \succeq_{\theta_i} r\vb) \right) \\
&\land & \exists \vb \in Q^*_i \textbf{ } \exists \vb' \in Q'^*_i \left(r(\vb') \succ_{\theta_i} r(\vb) \right).
\end{eqnarray*}
\end{definition}
Assessments in this space, as captured by preference relation $\succ_{\theta_i}$, must be sensitive to the entitlements of agents as represented by the relevant thresholds on $\vtheta_i \in E$.  The most significant benefits will be those that enable $i$ to function above one or more $\vtheta_i \in E$ when $i$ was not previously able to do so. The preference relation $\succ_{\theta_i}$ must, at a minimum, capture such improvements. Generally speaking, though, once an agent is able to function above all of the relevant thresholds on $\vtheta_i \in E$, then the impact of interactions is to be assessed relative to $M(Q_i, v_i)$ rather than $Q_i^*$.

\begin{remark}[Similarity between Condition~\ref{def:condition1} and Definition~\ref{def:assistance_objective}] Although both conditions range over $Q^*_i$, Definition~\ref{def:assistance_objective} characterizes a meritorious form of interaction (i.e., assistance through improvements to an agent's real freedoms), while Condition~\ref{def:condition1} states the minimum, necessary requirement for permissible interaction. Definition~\ref{def:assistance_objective} corresponds to improving $Q^*_i$ with respect to the preference relation $\succ_{\theta_i}$ while Condition~\ref{def:condition1} prohibits worsening $Q^*_i$---which is defined as making the set empty. Note that the latter is consistent with the minimal definition of $\succ_{\theta_i}$ presented above.
\end{remark}

\xhdr{Examples of AI assisting through improving real freedoms}
Assistive systems are often discussed in the context of disability, aging, sickness, injury or disease because these conditions threaten to bring $i$ below the relevant threshold $\theta(\vc_i)$ on some central capability.  But similar threats can be posed by natural disaster or humanitarian crises.  For example, earthquakes, hurricanes, tsunamis, and other natural disasters strip away or reduce critical infrastructure in a community (this would correspond to a negative impact on $\vs_i$), or hinder the ability of agents to access essential goods and services (negative impact on $\cX_i$).  This can make it a challenge for agent $i$ to function above the relevant threshold $\theta(\vc_i)$ on life, health, bodily integrity, affiliation, or other central capabilities.  In these cases, assistive systems perform tasks that preserve or expand the real freedom of agents by restoring elements of $\vc_i$, $\cX_i$, or $\vs_i$ to enable $i$ to function above $\theta(\vc_i)$.  This can include AI systems for identifying victims in collapsed structures, planning the most efficient and effective routes for supplies and aid. 

Alternatively, an interaction can be beneficial because of its impact on the agent's ability to advance the goals and ends that are constitutive of their conception of the good life.  More precisely: 

\begin{definition}[Assistance through Advancing Life Plans] \label{def:assistance_subjective}
    Consider an agent $i$ with initial freedom, $Q_i$.
    Let $Q'_i \neq Q_i$ denote $i$'s freedom \emph{following} an interaction. The interaction is \emph{beneficial/assistive through advancing $i$'s life plans} if
\begin{eqnarray*}
&& \forall \vb \in M(Q_i, v_i) \textbf{ } \exists \vb' \in Q'_i \left(v_i(\vb') \succeq_{\theta_i} v_i(\vb) \right) \\
&\land & \exists \vb \in M(Q_i, v_i) \textbf{ } \exists \vb' \in Q'_i \left(v_i(\vb') \succ_{\theta_i} v_i(\vb) \right).
\end{eqnarray*}
\end{definition}
In other words, the interaction either introduces a new way of enacting the agent's life plan without reducing other alternatives or reduces some of those alternatives but in return for introducing an alternative that the agent strictly prefers. 

\begin{remark}[Similarity between Condition~\ref{def:condition2} and Definition~\ref{def:assistance_subjective}] Although both conditions range over $M(Q_i, v_i)$, Definition~\ref{def:assistance_subjective} characterizes a meritorious form of interaction (i.e., assistance through advancing an agent's life plans) while Condition~\ref{def:condition2} states the minimum, necessary requirement for permissible interaction. Definition~\ref{def:assistance_subjective} corresponds to improving $M(Q_i, v_i)$ while Condition~\ref{def:condition2} prohibits worsening it (notice that $\succ$ in Definition~\ref{def:assistance_subjective} is replaced by $\succeq$ in Condition~\ref{def:condition2}).
\end{remark}

\xhdr{Examples of AI assisting through advancing life plans} AI systems that enable an agent to better advance goals that are constitutive of their conception of the good life, or that enable them to more effectively or efficiently carry out tasks that are instrumental to these goals, satisfy our second criteria for meaningful benefit and assistance.  For example, many small business owners cannot afford to build custom software for their business.  An AI system that facilitates entrepreneurial activities in low-resource communities can provide this type of assistance to an agent pursuing the success of their small business as a key component of their life plan. Generative AI models might provide a meaningful benefit to small business owners if they enable them to use plain language prompts to build web pages, marketing material, or to modify off-the-shelf business management software so that they can more efficiently run their enterprise.

Examples of this type of assistance can also parallel those mentioned under the previous form of assistance since disability, disease, and natural disaster or humanitarian crises are contexts in which agents face restrictions on, or risks to, their ability to realize opportunities critical to their set of maximal life plans.  In such cases, $i$ may be able to function above the relevant threshold $\theta(\vc_i)$ but not be able to function at the level necessary to take advantage of opportunities they value the most, as captured in their $M(Q_i, v_i)$. For example, an older adult facing the earliest stages of physical and cognitive decline may still be able to function above the relevant threshold $\theta(\vc_i)$ but nevertheless be at risk of not being able to realize opportunities in their $M$-set, because an alteration in $\vc_i$ given $\vs_i$ entails that $i$ is not able to convert $\vx_i$ into their preferred $\vb_i$ $\in M(Q_i, v_i)$. In this case, an assistive system is one that will close the gap between $\vc_i$ and $\vx_i$ in $\vs_i$ to enable $i$ to realize their choice of opportunities in their $M$-set.  
For example, cooking might be an activity that $i$ enjoys as an outlet for creativity and pleasure or it might be something that $i$ needs to accomplish to remain independent but that is not constitutive of a good life.  In either case, as $i$ begins to experience cognitive decline it becomes more difficult to manage their grocery shopping---to plan out meals, purchase the relevant ingredients, and avoid purchasing multiple instances of the same items.  This can make it more difficult to live within their budget, to maintain their health, to enjoy meals and to entertain.  An AI system that could help $i$ create an appetizing and healthy meal plan that fits within their budget, ensure that they have the requisite items on hand, and even help $i$ with the steps of the meal preparation process would support functions that $i$ requires to maintain their independence. 

It is worth emphasizing that, in all such cases, the real freedom of agents and their ability to actualize elements of their life plan are mediated by the norms and infrastructure of the society in which they live, that is, $\vs_i$.  As a result, it will always be an ethical question whether the best way to support an agent is by adding resources to their resource base, $\cX_i$, trying to improve their relevant characteristics, $\vc_i$, reforming social norms or improving the infrastructure, $\vs_i$, or some combination of these.

It is natural to think of high-stakes contexts as circumstances in which $i$ faces a threat to or a restriction on their real freedom or their ability to take advantage of opportunities in their $M$-set.  Sickness, injury, disease, disability, aging, and natural disasters or humanitarian crises seem like paradigm cases of high-stakes contexts.  It is worth noting, therefore, that when individuals come to rely on an AI system to carry out functions or tasks that are critical to their real freedom or their ability to realize opportunities in their $M$-set, this dependency \emph{creates} a high-stakes context of interaction.  The reason is simply that $i$'s reliance on that system in these ways makes them particularly vulnerable to faults, flaws, or errors with that system. Their real freedom and their well-being are dependent on the ability of those systems to function.  As AI systems become more capable and proliferate across various aspects of our lives, our dependence on them grows and the importance of their functioning as assistive systems increases.

\subsubsection{Assistance through Justified Paternalism}\label{sec:justified_paternalism}

Ideally, assistive systems empower agents by expanding their ability to do or to be in ways that advance their conception of the good life.  In this way, the goals of the AI system---understood as the set of tasks the system has been trained to perform---and of the agent are aligned.  Conflict between the goals of an AI system and the considered goals of the agent is a source of ethical tension.  However, there are cases where conflicts 
between the agent's apparent goals and the goals of the AI system can be morally beneficial. Understanding these cases is important because it allows us to demonstrate how our model can capture clear criteria for distinguishing cases of permissible interference with the agent's goals from impermissible cases. Because the difference between these cases hinges on the extent to which the interference with another's agency is grounded in an understanding of and a responsiveness to their considered life plan, it also reinforces the value of having a model that can represent the relationship between an agent's life plan and their well-being. 

Consider the following case adapted from \citep{goldman1980moral}.  Two friends, $i$ and $j$, are talking on a subway platform and $j$ knows that $i$ wants to take the A train uptown.  But $i$ is so wrapped up in conversation that when the downtown train arrives $i$ proceeds to say goodbye and to jump onto the train.  Noticing $i$'s error, $j$ steps onto the train, grabs $i$ by the hand and drags them off of the train.  $i$ is noticeably upset until $j$ points out that $i$ was on the wrong train. 

In the above example, $j$ limits $i$'s freedom by choosing to make actual a particular $\vb_i$ when $i$ seemed to prefer $\vb'_i$ through their choice. But note that $i$'s default choice in this example is not well-informed and considered, due to $i$'s distraction with the conversation. This is an example of paternalistic action because when $j$ acts, $j$ chooses to make actual for $i$ the $\vb_i$ that $j$ regards as best for $i$, given $i$'s well-informed and considered values.  When the subject of the paternalistic action is an otherwise competent adult, paternalistic action is regarded as unethical unless it can be justified.  Paternalistic actions are justified under the following conditions\footnote{Conditions a--c are drawn from \citep{goldman1980moral} pp.162--166.  We add condition d for completeness.}: 
\begin{enumerate}[label=(\alph*)]
    \item $j$ knows what $i$ would choose, had $i$ been capable of making a well-informed, considered choice given their goals and values, represented by $v_i$;
    \item $i$ is ignorant of some fact or facts that, were they to know, would lead them to choose $\vb_i$ given $v_i$; 
    \item it isn't possible under the circumstances \footnote{This condition could be met for several reasons.  For example, $j$ might simply not have time to inform $i$ of the relevant information as in the example provided.} to inform $i$ of this information so that they can choose $\vb_i$ for themselves, and 
    \item the means used to bring about $\vb_i$ satisfy a proportionality condition in that any burdens they impose on $i$ do not outweigh the risks associated with $i$ missing out on $\vb_i$. 
\end{enumerate} 
In our example, $j$ is justified in grabbing $i$ and hauling them off of the train since the conditions above are met. But $j$ would violate the proportionality condition if, for example, they stopped the entire train from leaving the station or if they had injured $i$ so that they could not board the incorrect train.

Cases of justified paternalism are likely relatively rare.  Nevertheless, in clear cases where paternalism is actually justified, $j$'s conduct constitutes a form of assistance. 

This is because $i$'s conduct has become misaligned with $i$'s own goals and values and $j$ knows this.  Understanding this purely conceptual point is important because it provides confirmation that the focus of assistance to $i$ should be defined relative to the plans and projects that $i$ values most. In other words, the fact that paternalism is regarded as justified when $j$ acts so as to ensure that $i$ chooses in accordance with $i$'s own goals and values provides a socially accepted example in which the moral status of $j$'s conduct is determined by its relationship to the values and goals that constitute $i$'s life plan.

Similar confirmation comes from even more extreme cases in which $i$ lacks the ability to make decisions for themselves. In such cases it is widely recognized that proxy decision makers should attempt to make the decision for $i$ that $i$ would make, given their expressed goals and values, if they were able to understand their circumstances, the available options, and to bring their values to bear to make a choice.  This is sometimes known as the ``substituted judgment standard''\footnote{ The substituted judgment standard is sometimes challenged on epistemic grounds because it can be difficult for proxies to know what an agent would want in specific circumstances.  See, for example \citep{dresser1989quality}.  Among other replies in defense of this standard, see \citep{rhoden1989limits, kestigian2016adversaries}.} or the ``subjective standard of proxy decision making'' and it dates back as early as 1860 in English law (Strunk v. Strunk, 445 S.W.2d 145, 148 (Ky. 1969)); see also in re Quinlan, in re Conroy, the Uniform Guardianship, Conservatorship, and Other Protective Arrangements Act. 
Widespread support for this standard reflects key insights on which our model is built, namely, that in high-stakes contexts decisions that impact $i$'s interests should be made relative to the goals and values (i.e., $v_i$) that shape $i$'s conception of the good life, $M(Q_i, v_i)$.

We emphasize that the above conceptual point should not be mistaken for the claim that AI systems should be designed to act paternalistically.  Generally speaking, paternalistic interactions should be minimized wherever possible.  First, limiting the agency of a person is prima facie morally problematic and should be avoided wherever possible.  Second, $j$'s certainty in their estimate of $v_i$ and consequently, $M(Q_i, v_i)$ is likely not to be well-calibrated, leading to cases in which $i$'s freedom is restricted out of an incorrect conception of $i$'s own good.  Such mistakes are likely to be common and to result in interactions that constitute unjustified restrictions on $i$'s freedom and that fail to advance $i$'s well-being.  We return to these issues in Section~\ref{sec:failure_modes} where we define unjustified paternalism.

\xhdr{Examples of justified paternalism through AI}
Having said that paternalism should be avoided whenever possible, there are cases in which the assistance that AI systems provide to individuals can take the form of justified paternalism.  For example, $i$ might know that, relative to $v_i$, they do not want to send certain texts or emails but that, if they become drunk, they are likely to want to do that (i.e., in a drunken state they are likely to see this as the best option relative to their transient $u_i$ and to disregard $v_i$). Before $i$ goes drinking, $i$ might install software on their computer to prohibit such activity.  Current versions of this rely on some proxy for inebriation, such as the agent's inability to solve relatively simple but still detailed math problems.  One can imagine more sophisticated examples of such a system.  Central to the permissibility of such systems is the fact that they are initiated by the agent themselves, to avert specific behavior they can anticipate, where the restrictions on their own ability to function are limited in both duration and extent.  One can also imagine systems with a similar functionality but in different domains.  For example, scams that target older adults function through emotionally compelling promises of wealth or by giving the impression that a loved one is in need.  In a cool hour, reflecting on one's vulnerability to such ploys, an agent might deploy an AI system that would disconnect scam phone calls, delete scam texts, or prevent the purchase of gift cards after the receipt of such messages. Similar applications may limit the time the agent spends on social media platforms. Additional examples of justifiably paternalistic AI systems can be found in the context of semi-autonomous driving, where the vehicle may utilize visual or haptic cues or even automatically brake when the human driver is distracted.

In summary, this Section established the criteria for justified paternalism, showed that those criteria support a focus on the life plan of the agent when understanding assistance, and that, although paternalism should be avoided where possible, there can be use cases where these criteria are met by AI systems that act on these terms to provide assistance to individuals.

\section{Notable Failure Modes}\label{sec:failure_modes}

In the previous section, we argued that assistive interactions are morally meritorious because they expand or improve an agent's real freedom or their ability to pursue their preferred life plan. They often express a relationship of respect because they promote agent $i$'s ability to choose a $\vb_i$ from $M(Q_i,v_i)$.  Interactions that deviate from these criteria can alter the relationship between an agent's capabilities, freedom, and well-being in ways that are morally problematic. It is helpful, therefore, to contrast assistive interactions with unjustified paternalism, coercion, deception, exploitation and domination. We regard each of these forms of interaction as an independent moral wrong. Although there is scholarly disagreement about how each of these wrongs should be defined, how their boundaries should be drawn, and whether their wrongness can be explained in terms of more fundamental moral concepts, the model we have presented here can explain the moral wrongness of these interactions in a significant number of cases.
Formally, while our second form of assistance (i.e., Definition~\ref{def:assistance_subjective}) consists of expanding or improving the set of maximal achievable life plans, $M(Q_i, v_i)$, for agent $i$, the morally problematic interactions outlined in this section shrink or worsen this set for the agent, forcing them to choose their functioning in accordance with the goals of another agent, say $j$. As a result, many of these interactions will violate Condition~\ref{def:condition2}.  In particularly serious cases they will violate Condition~\ref{def:condition1} because the interaction will drop $i$ below some component of $\vtheta_i$, e.g., on life, bodily health, bodily integrity, practical reason or affiliation.

\begin{figure}[t!]
    \centering
    \includegraphics[width=0.6\textwidth]{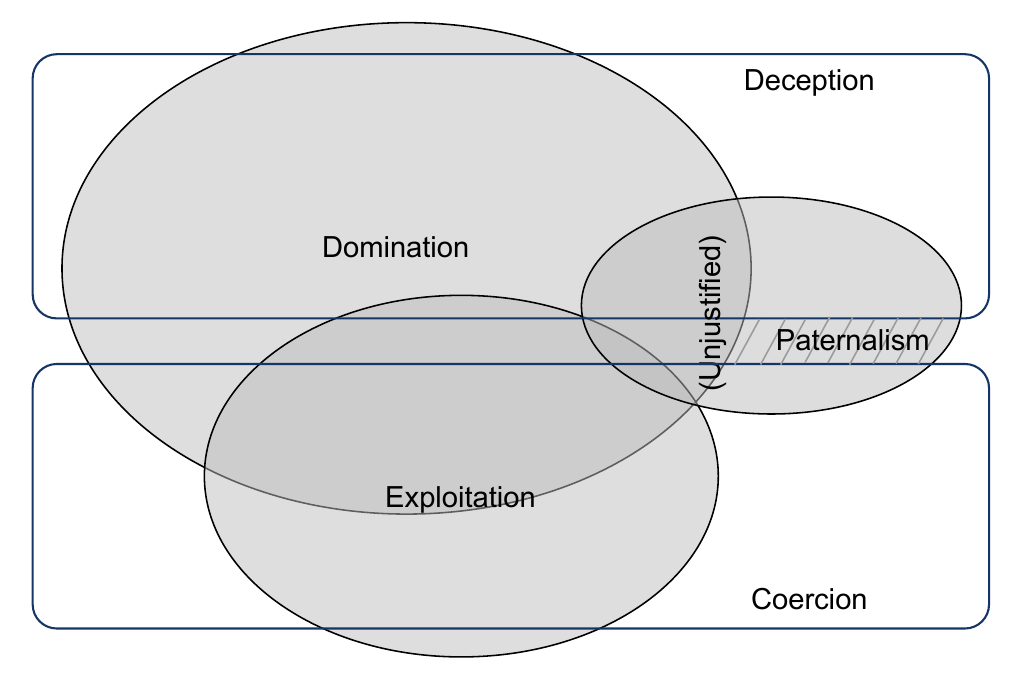}
    \caption{A simplified Venn diagram illustrating the relationship between domination, exploitation, paternalism, coercion, and deception.}
    \label{fig:failures}
\end{figure}

We begin with unjustified paternalism, in which the freedom of $i$ is limited out of a concern for $i$'s well-being but the conditions for justified paternalism are not met.  Since paternalism is often facilitated through coercion or deception, we explicate those relationships next. We then turn to exploitative interactions, which can, but need not, involve coercion or deception.  We conclude with domination, which is a general species of undue influence under which both unjustified paternalism and exploitation fall. Figure~\ref{fig:failures} illustrates the high-level relationship between the failure modes enumerated in this section.

\subsection{Unjustified Paternalism} 
Given the discussion in Section~\ref{sec:justified_paternalism}, unjustified paternalism is one of the clearest examples of an interaction misaligned with the agent $i$'s life plan.  This occurs when another agent, $j$, makes available for choice to $i$ a set of functioning vectors $Q'_i$ that $j$ regards as best for $i$, but the justificatory burden necessary to sanction modifying $Q_i$ to $Q'_i$ hasn't been met.  In other words, unjustified paternalism occurs where: 
\begin{enumerate}[label=(\alph*)]
    \item $j$ has misjudged $i$'s goals and values, so acts in a way that deviates from what $i$ would value and choose for themselves, or
    \item $i$ acts not in ignorance but according to their well-informed and considered values, or
    \item circumstances allow $j$ to communicate the relevant necessary information to $i$ so $i$ can make an informed choice, or
    \item the means $j$ uses to bring about new functioning vectors for $i$ (say $\vb_i$) imposes greater burdens on $i$ compared to the risks associated with $i$ missing out on them (here, $\vb_i$).
\end{enumerate}

In cases where the choices $i$ faces relate directly to their ability to function in ways that they value as constitutive of their preferred life plans or that are instrumental to those goals, unjustified paternalism violates Condition~\ref{def:condition2}.  In cases where the choices $i$ faces relate to their basic capabilities, e.g., to express their sexuality, control their body, safeguard their health, or affiliate with others, instances of unjustified paternalism violate condition~\ref{def:condition1}. The only cases that would not constitute a violation of one of these conditions would involve interactions in which $j$ restricts $Q_i$ only relative to a choice that is insignificant in that it does not alter $i$'s real freedoms or $i$'s choice of functionings that are constitutive of, or instrumental to, their conception of the good life.

\xhdr{Examples of unjustified paternalism through AI} 
Cases of unjustified paternalism can occur when $j$ purchases or designs an AI product with the goal of providing assistance to $i$ without understanding the goals and values (i.e., $v_i$) that define $i$'s conception of the good life ($M(Q_i, v_i)$). They can also arise when there is a value conflict between the goals that $j$ wants to advance for $i$ and the goals that $i$ most values.  For example, $j$ might be concerned about the health and safety of $i$ and so purchase a system that monitors and then reports information about key events back to $j$. Here $j$ values safety for $i$ over privacy for $i$.  But if $i$ values privacy over safety---being able to pursue certain activities without $j$ awareness---then the system is advancing $j$'s conception of $i$'s best interests rather than $i$'s.

Acts of unjustified paternalism often involve, as mechanisms, \emph{coercion} and \emph{deception}.
One reason that $i$ might acquiesce to $j$ in cases of unjustified paternalism is that $i$ depends on $j$ in ways that affect $i$'s ability to realize opportunities in their $M$-set.  Leveraging such dependencies can constitute \emph{coercion}. Alternatively, $j$ might be able to bring $i$ to choose $j$'s preferred functioning vector for $i$ if $j$ can exert sufficient control over the information available to $i$.  Filtering the information available to $i$ in this way constitutes \emph{deception}.  Because coercion and deception can also be mechanisms for domination and exploitation, we explicate the former before returning to the latter.

\subsection{Coercion} 
In a coercive interaction, $j$ brings $i$ to choose from a restricted set of $\vb_i$'s through the use of force or by threatening to make $i$ worse off if $i$ chooses any other $\vb'_i$ not in the set~\citep{Nozick1969Coercion}. In other words, $j$ threatens to change $i$'s freedom to act, i.e., $Q_i$, by producing a new $Q'_i$ that $i$ regards as worse on some dimension than their initial $Q_i$ where $j$ does not have an independent right \footnote{Although we typically use the term ``coercion'' to pick out moral violations, there are cases in which coercion can be justified.  In the criminal justice context, for example, prisons exert coercive force over inmates. If those institutions are in line with principles of justice, then the coercive restraints they place on inmates may be morally justified.  In those cases, prison officials may coerce inmates, but, because they have a prior right to impose such restrictions, their actions would not be morally wrong.} or prerogative to impose such a restriction on $i$ \citep{wertheimer1987Coercion}. Because threats are only effective when they implicate something an agent values, coercion often involves a threat to make $i$ worse off relative to $M(Q_i, v_i)$.  By implication, in such cases coercion will almost always constitute a violation of Condition~\ref{def:condition2} and likely a violation of Condition~\ref{def:condition1} since effective threats of harm are likely to involve actions that would drop $i$ below some $\vtheta$, e.g., on life, bodily health, or bodily integrity. While threats to make the agent worse off with respect to $v_i$ or $r$ characterize most \emph{serious} forms of coercion, less serious kinds of coercive interactions are possible to the extent that $j$ tries to influence $i$'s choice of functionings with respect to $u_i$\footnote{This point can be made more broadly for all the problematic interactions outlined in this Section. The most serious failures are those that are induced with respect to $v_i$ or $r$, but they can still be conducted with respect to the agent's transient/mundane valuation, $u_i$.}, i.e., over matters that are simply unimportant to $i$, such as where to eat, or where to sit at some particular event.   

\xhdr{Examples of coercion through AI}
Coercion can be accomplished through explicit and implicit threats.  In the above example of unjustified paternalism (i.e., unwanted surveillance of an older adult), for example, $i$ might know that if $i$ does not choose the $\vb'_i$ favored by $j$, $j$ will withhold or otherwise reduce the resources on which $i$ depends (i.e., change $\vx_i$ to a strictly worse $\vx'_i$), refuse to assist $i$ with some functioning that $i$ values or physically abuse $i$ (i.e., change $\vc_i$ to a strictly worse $\vc'_i$). 
In this example, $j$ uses coercion to create the context in which $i$ is willing to use or be subjected to the use of an AI system.  Beyond this example, AI systems might be designed by developers or deployed in ways that exert coercive force on agents.   Ransom-ware, for example, unilaterally prevents an agent $i$ from accessing their data or system functions unless $i$ pays a fee to a third party.  In this case, software carries out the function that $j$ uses to bring $i$ to choose the $\vb'_i$ favored by $j$ when it restricts $i$'s freedom to act, $Q_i$, by producing a new $Q'_i$ that $i$ regards as worse on some dimension than their initial $Q_i$ and $j$ does not have an independent right or prerogative to impose such a restriction on $i$.

As more social transactions are carried out online, certain services can become essential gateways to participating in those transactions.  As dependency on those gateways increases, firms that control them can leverage their importance to extract concessions from users, e.g., related to personal data privacy, using the threat of locking them out of essential services.  As a concrete example, as job openings migrate to online systems, firms could exert coercive force on users by requiring access to a wider range of sensitive, personal, or private information in return for access to openings and placement opportunities.  

\subsection{Deception} 
In a deceptive interaction, $j$ gets $i$ to choose one of $j$'s preferred functioning vectors, say $\vb'_i$, rather than those preferred by $i$, say $\vb_i$, by shaping the information available to $i$.  For example, $j$ might ensure that $i$ is not aware that $\vb_i$ is available for choice, or might misrepresent the feasibility of $i$ choosing $\vb_i$ by misrepresenting the resources, $\vx_i$, available to $i$, or mischaracterizing $i$'s ability, $\vc_i$, to convert existing resources into their choice of functionings, through the choice of the relevant pattern of use, $f$. Alternatively, $j$ might misrepresent the relative value of choosing $\vb_i$---the functioning vector $i$ would have chosen without $j$'s deceptive interference--so that it appears to be dominated by $\vb'_i$.  Here again, deception will constitute a violation of Condition~\ref{def:condition2} to the extent that it reduces $i$'s opportunity to advance ends that are constitutive of, or instrumental to, $i$'s conception of the good life and violates Condition~\ref{def:condition1} to the extent that it reduces opportunities in $Q_i^*$ necessary for $i$ to function above $\vtheta$ on a relevant dimension such as bodily integrity, control over one's environment, or practical reason. 

\xhdr{Examples of deception through AI} 
In the example from unjustified paternalism (i.e., unwanted surveillance of an older adult), the older adult $i$ might use an AI system because $j$ hides the true extent of the information the system records or tracks.  More generally, creators of AI systems can take steps to obscure the way their systems perform, e.g., how they gather and processes information, to prevent agents from leaving and addressing their problem through another means.  As AI systems become capable of interacting with individuals in increasingly complex ways, they can become instruments for facilitating deceptive interactions.  For example, as currently designed, certain large language models can encourage users to perceive the AI system as sentient, as having feelings, or as having psycho-social capabilities they do not possess.  This can lead users to disclose sensitive personal information to the model, to seek counseling or advice, and to act on model outputs that conflict with the agent's well-informed interests.  The ability of these models to influence human emotions and judgments is exacerbated by their tendency to speak with authority, or to package outputs as conveying emotional valence, such as compassion or empathy, even when they information they present is false or misleading. 

Having explicated the concepts of coercion and deception, next we distinguish paternalism from two other forms of misaligned interaction, namely \emph{exploitation} and \emph{domination}.  

\subsection{Exploitation} Exploitation and paternalism are related in that the agent $j$ who is exploiting $i$ or acting paternalistically toward $i$ acts so that one of the functioning vectors that $j$ regards as best is realized for $i$. Call this functioning vector $\vb_i$. In the case of paternalism, $j$ regards $\vb_i$ as best relative to $j$'s conception of $i$'s well-being.  In exploitation, in contrast, $j$ regards $\vb_i$ as best relative to either $j$'s material advantage or to the material advantage of some third party, without regard to the relationship of $\vb_i$ to $M(Q_i, v_i)$.  More generally, both paternalism and exploitation involve influencing the choices or behavior of $i$ but where the goal in paternalism is to benefit $i$, the goal of exploitation is to create material benefit or value for someone other than $i$. 

There is some disagreement in the literature about how exploitation should be understood.  For example, some authors note that exploitative interactions often involve coercion, deception or some combination of the two \citep{holmstrom1977exploitation, schwartz1995s}.  In those cases, exploitation will almost always constitute a violation of Condition~\ref{def:condition2} and will likely violate Condition~\ref{def:condition1} for the reasons already articulated in the discussion of coercion and deception. 
Other scholars note that, although exploitation can involve unwarranted interference with $i$'s agency, this need not always be the case \citep{sample2003exploitation, wertheimer1999exploitation}.  Exploitative offers might introduce a new option into $Q_i$ and $i$ might freely accept that offer, with full information about its consequences, but the interaction may nevertheless be impermissible because it violates an independent standard of fairness. For Wortheimer, that standard is set by something like the division of the surplus that would be generated in an ideal market \citep{wertheimer1999exploitation}.  However, others have argued that when $i$ is badly off then the relevant standard of fairness should be determined by a baseline in which the worse off party receives the largest share of the surplus \citep{ballantyne2010research}.  We follow \citep{nussbaum1999women} in holding that an agent's ability to function above the minimum acceptable threshold on the capacity for affiliation requires being treated as the moral equal of others and as a being with dignity--as an agent rather than as an instrument purely for the benefit of others.  This includes being treated in accordance with the appropriate standard of fairness in economic and social transactions.  As such, fairness-based accounts of exploitation can be explained, in our model, as a violation of our Condition~\ref{def:condition1}.

\xhdr{Examples of exploitation through AI} 
As the capabilities of AI systems increase, so does the prospect that they will be used for exploitative purposes.  For example, ransom-ware is used to coerce $i$ into choosing the $\vb_i$ that is preferred by $j$, paying money to $j$ so that $j$ does not delete $i$'s data or render $i$'s system useless where $j$ regards this option as best because it generates profits for $j$ or their accomplices.  Likewise, ``dark patterns'' can involve deceptive practices in which $i$ engages with a system in the hope of choosing $\vb_i$ that is best according to $v_i$ but is brought to select some $\vb'_i$ that is worse than $\vb_i$ but that advances $j$'s interest \citep{mathur2019dark,mathur2021makes}.  However, dark patterns need not involve outright deception, understood as the assertion or communication of information that is false or misleading.  They might simply target aspects of $i$'s choice architecture that encourage $i$ to choose $\vb'_i$ rather than the $\vb_i$ that is best according to $v_i$ \citep{narayanan2020dark}. 

In contrast, concerns about exploitative interactions not involving deception or coercion have also been raised in regard to the use of AI. Concerns of this kind are rightly raised about the commodification of attention and the extent to which social media platforms benefit from user engagement, even if the nature or duration of the engagement has a negative impact on the health or well-being of users \citep{bhargava2021ethics, castro2020attention, rubel2021algorithms}.  Similarly, questions have been raised about the extent to which AI systems rely on user generated data to improve their performance (see, e.g.,~\citep{openAI_performance}) and subsequent revenue for creators and owners without offsetting compensation to users.  More broadly, the reliance of many AI systems on the labor of poorly compensated workers, often from low or middle-income countries has also raised concerns about exploitation in the development and refinement of AI systems~\citep{openAI_kenya}.

\subsection{Domination} 
Each of the relationships and interactions listed in this section differs from assistance in one key respect: $i$ chooses or performs some $\vb_i$, not because it is best relative to the goals and values ($v_i$) that defines their conception of the good life, $M(Q_i, v_i)$, but because it is best relative to the goals and values of some other agent, $j$.  In this regard, the failure modes discussed above can each be seen as a different species or form of domination.  In a relationship of domination, the dominating agent $j$ has sufficient power over the information available to $i$, the circumstances in which $i$ chooses, or some other factor that might modulate $i$'s choice such that when $j$ desires that $i$ choose to realize some $\vb'_i$, $i$ chooses to realize that $\vb'_i$, regardless of how $\vb'_i$ relates to $v_i$.  Domination is similar in form to exploitation, in that $j$ brings $i$ to choose the $\vb'_i$ that $j$ most values, but relationships of domination can advance goals of $j$ that range beyond the narrower set of material benefits usually associated with exploitation.

Domination is the antithesis of assistance because $j$ uses the powers at their disposal to shape $i$'s conduct so that it advances $j$'s goals without regard for its relationship to $M(Q_i, v_i)$.  In relationships of assistance, $j$ seeks to support $i$'s ability to choose some $\vb_i$ because of its importance to $i$'s conception of the good life, $M(Q_i, v_i)$.  Assistance thus embodies profound respect for $i$ as a person---as the subject of a life which they lead and experience and in whose quality and character they are deeply invested. In relationships of assistance, $i$ and $j$ pursue the same goals because $j$'s concern for $i$ aligns $j$'s interest with $i$'s---$j$ chooses to assist $i$ because enabling $i$ to advance $M(Q_i, v_i)$ has value for $j$.  In contrast, domination embodies profound disrespect for $i$---reducing $i$ to an instrument $j$ uses to advance $j$'s ends without regard for $i$ as a subject of their own life. 

As with the other failure modes mentioned in this section, domination often involves a violation of Condition~\ref{def:condition2} whenever $j$'s power and influence is used to alter the choice of $\vb_i$ from $M(Q_i, v_i)$ and will violate Condition~\ref{def:condition1} in cases where the choices $i$ faces relate to their ability to function above the minimum acceptable thresholds, captured in $\vtheta$, on some basic capability in $Q_i^*$. Examples include the agent expressing their sexuality, controlling their body, safeguarding their health, affiliating with others. The only cases that would not constitute a violation of one of these conditions would involve interactions in which $j$ restricts $Q_i$ only relative to a choice that is insignificant in that it does not alter $i$'s real freedoms or $i$'s choice of functionings that are constitutive of, or instrumental to, their conception of the good life.

\xhdr{Examples of domination through AI} 
As an example, consider the deliberate spread of misinformation during political campaigns. If $j$ can determine which information a social media platform displays to users, $j$ might influence the social sentiments of users to support broad social or political movements.  Promoting an ideology can be motivated by monetary or other direct forms of benefit, but need not be.  In relationships of domination, the concern is less about the distribution of the resulting benefit and more centrally about the legitimacy of the influence that $j$ exerts over $i$'s choices and decisions.

As AI systems become capable of mediating more complex relationships between people, our dependence on those systems renders us vulnerable to various forms of domination.  It thus becomes ever more important to ensure that such systems are designed and used in ways that empower people to advance their well-being.

\section{Conclusion}\label{sec:discussion}

The increasing sophistication and capability of AI systems underscores the importance of ensuring that when they interact with humans, they do so in ways that are, at the very least, morally permissible and preferably meaningfully beneficial.  The formal model outlined here captures important necessary conditions for moral permissibility, shows how beneficence is connected to concepts such as basic rights, various senses of moral freedom, and individual well-being, and connects these concepts to a finer taxonomy of morally problematic interactions (including unjustified paternalism, coercion, deception, exploitation, and domination) expanding the lexicon of morally salient issues to consider in conversations about responsible and ethical AI. 

The model that we articulate here also highlights how the increasing ability of AI systems to impact the freedom and well-being of individuals elevates the importance of ensuring that such systems are designed, from their inception, in ways that align their functioning with the rights and well-being of individual stakeholders. In particular,  within the model we present \emph{value alignment} can be reconceptualized as ensuring that AI systems operate in conformity with our necessary conditions on morally permissible interactions and that they provide some form of assistance to individuals impacted by these interactions.  It also consists in ensuring that agents rely only on systems capable of operating in ways that show respect for their standing as moral agents by advancing their real freedom or their well-being. Misalignment, in contrast, involves cases where agents are induced to rely on, and therefore become vulnerable to, AI systems that cannot or do not respect their status as agents, worsen their real freedoms, or impair their ability to advance their considered life plans.

This new conception of AI alignment has several virtues.  First, it forges a needed connection between the values that must govern the design and use of AI systems and the impact of the systems on agents. Second, it creates a bridge between immediate concerns about the individual and social impacts of current AI systems and longer-term impacts associated with the development of future AI systems.  In particular, ensuring that AI systems respect the necessary conditions for morally permissible interactions outlined here provides a strong social safeguard against both concrete, immediate, harms and more distant but catastrophic possibilities.  AI systems that function in ways that respect the status of individuals as moral agents, by safeguarding their basic interests and respecting or promoting their ability to advance their considered life plan (consistent with the basic interests of others in being able to do the same) will be less likely to harm individuals or to set back their interests in minor or major ways.  Cultivating beneficent AI systems is thus an attractive way to promote AI ethics and AI safety, simultaneously.

Our work suggests several important avenues for future work:

\xhdr{Imagining and evaluating possible AI designs through the lens of our framework} The conceptual framework presented here has the potential to guide creators of AI systems in (a) assessing the ethical ramifications of their work according to a wider set of criteria beyond the common principles of responsible AI, and (b) imagining new possibilities for the design of assistive AI systems---by centering users' life plans and basic rights. It can, additionally, serve as a new foundation for participatory approaches that elicit what functionings of AI designs could meaningfully assist the target stakeholder groups, such as capability sensitive design \cite{jacobs2020capability}. We leave a careful exploration of these ideas for future empirical work.

\xhdr{From benefiting individuals to larger groups and communities} The basic version of the model presented here conceptualizes assistance at the level of \emph{individual} agents. However, the framework can be readily expanded to reason about assistance to larger populations through mass market AI products. In such cases, one may define assistance with respect to common/average life plans of the target user group. We note, however, that in such cases it is critical to consider several additional risks: it is possible for an AI product to be assistive to a large number of individuals but target the wrong group of people (e.g., the already privileged at the expense of the historically under-served and underprivileged.) These justice-related considerations are critical to address in the near future. 

\xhdr{Conceptual connections to justice and fairness}
Our model, as articulated here, does not address the question of how to prioritize the interests of the various stakeholders who might benefit from the development of AI systems.  This reaches beyond the question of what constitutes beneficence and assistance and into the realm of justice since it requires an assessment of who has a social claim to the benefits that might be produced from assistive AI systems and how those claims might be ordered.  
Finally, it is worth emphasizing that the analysis presented here focuses specifically on a closely related cluster of moral concepts: beneficence, freedom, and respect.  Because these concepts are so fundamental, the model outlined here is likely to have important implications for aspects of AI ethics and responsible AI that presuppose or build on these concepts.  This includes issues of equity and fairness insofar as these concepts deal with both respect for the equal status of persons and the distribution of the benefits and burdens of AI systems.  Exploring these connections is an important element of future work.

\bibliographystyle{apalike}
\bibliography{main}

\end{document}